\documentclass[journal]{IEEEtran}
\ifCLASSINFOpdf
\else 
\fi
\usepackage{cite}
\usepackage{csquotes}
\usepackage[english]{babel}
\usepackage{subfig}
\usepackage{graphicx}
\usepackage{flushend} 
\usepackage[skip=0pt]{caption}
 \usepackage[figurename=Fig.]{caption}
\usepackage{float}
\usepackage{amsmath}
\usepackage{algorithm}
\usepackage[noend]{algpseudocode}
\usepackage{accents}
\usepackage{tabularx,booktabs}
\newcolumntype{C}{>{\centering\arraybackslash}X}
\usepackage{makecell}
\usepackage{multirow}
\usepackage{tabularray}
\usepackage{nomencl}
\usepackage{xcolor}
\makenomenclature

\newcommand\munderbar[1]{
  \underaccent{\bar}{#1}}
\begin{document}
\title{AT-2FF: Adaptive Type-2 Fuzzy Filter for De-noising Images Corrupted with Salt-and-Pepper}

\author{Vikas Singh \IEEEmembership{Member,~IEEE}
\thanks{Vikas Singh is with the Dept. of
School of Life Science, Gwangju Institute of Science and Technology,
South Korea (e-mail:  vikkysingh07@gmail.com)}
}

\maketitle

\begin{abstract}
Noise is inevitably common in digital images, leading to visual image deterioration. Therefore, a suitable filtering method is required to lessen the noise while preserving the image features (edges, corners, etc.). This paper presents the efficient type-2 fuzzy weighted mean filter with an adaptive threshold to remove the SAP noise. The present filter has two primary steps: The first stage categorizes images as lightly, medium, and heavily corrupted based on an adaptive threshold by comparing the M-ALD of processed pixels with the upper and lower MF  of the type-2 fuzzy identifier. The second stage eliminates corrupted pixels by computing the appropriate weight using GMF with the mean and variance of the uncorrupted pixels in the filter window. Simulation results vividly show that the obtained denoised images preserve image features, i.e., edges, corners, and other sharp structures, compared with different filtering methods. The code and experimented data of the AT-2FF is available on the GitHub platform: {https://github.com/vikkyak/Image-Denoising}. \\

\end{abstract}

\IEEEpeerreviewmaketitle
\section{Introduction}
\label{sec: Introduction}

\IEEEPARstart{I}{n} the modern image processing system, image denoising plays a vital role in suppressing the noise from the image. The noise adversely affects the image processing task,  image acquisition, video processing, image analysis, coding, tracking, etc. Impulse noise inevitably influences the natural image during image acquisition, coding, transmission, and imperfection in sensors. Removing and restoring the original image to speed up the image processing task, such as image detection, segmentation, object recognition, etc.  \cite{ce}.

In image processing, the primary purpose of image denoising is to preserve the edges, corners, and detailed image information. It has been studied in the literature that non-linear filtering methods perform better than linear filtering methods \cite{lf}. The non-linear filtering methods, including standard median filter (MF) \cite{MF}, weighted median filter (WM)  \cite{wm}, adaptive median filter (AMF) \cite{amf},  adaptive weighted median filter (AWMF) \cite{AWMF}, and truncated median filter (TMF) \cite{TMF}, have been used for filtering impulse noise. However, these filters blur the images and do not preserve edges while removing impulse noise at higher values. In \cite{mamf}, a modified AMF is presented in which the median filter replaces the corrupted pixels or their neighbor's pixel values estimate the corrupted pixels. In \cite {namf}, a weighted mean filter is utilized to lessen the SAP noise by assuming equal weights of uncorrupted pixels.

 The past success of fuzzy filters due to an extra degree of fuzziness provides a more helpful way of handling uncertainty in a noisy environment.   In the literature, weighted fuzzy mean (WFM)\cite{c6}, adaptive fuzzy (AF) \cite{c7}, fuzzy c-means clustering filter (FRFCM) \cite{FRFCM}, iterative adaptive fuzzy (IAF) \cite{c8} and adaptive fuzzy switching weighted mean \cite{c9} filters are presented to reduce salt and pepper (SAP) noise. The filtering output of WFM is obtained by membership value with their associated fuzzy rule base. In contrast, the AF filter works on filter window enlargement to identify and denoise SAP noise. These filters still have a problem with weight assignment and cannot preserve detailed image information. 

Recent studies have suggested that the type-2 fuzzy filter can figure them out more efficiently with improved performance due to uncertainty modeling \cite{c25}. Liang and Mendel have proposed a fuzzy filter with a type-2 Takagi-Sugeno-Kang (TSK) fuzzy model to remove the SAP noise \cite {c13}. Yıldırım \emph{et~al.} \cite{c14} have proposed a detailed fuzzy filter based on type-2 fuzzy and experimented on different noise density levels. The problem with these methods is the high computational complexity for a large rule base. To overcome the rule-based problem recently, authors have presented type-2 fuzzy filter \cite{singha, singhi} and type-2 fuzzy filter with Matrix Completion (T-2FFMC) \cite{T-2FFMC}. These methods perform well, but choosing a reasonable threshold is still problematic. In \cite{h1}, authors have presented a two-stage adaptive filter with fuzzy weight switching to remove the SAP noise. The problem with this method is that the threshold is heuristic and does not preserve the image features at higher noise. In the present work, I have presented a two-stage fuzzy filter: The first stage categorizes images as lightly, medium, and heavily corrupted based on an adaptive threshold using a type-2 fuzzy approach with maximum absolute luminance difference (M-ALD) of the processed pixels, and the second stage eliminates corrupted pixels by computing the appropriate weight using GMF of the uncorrupted pixels in the filter window. The simulation results show that the present approach preserves the image details at higher and lower noise levels.

The notable contributions of the paper are summarized as:
\begin{enumerate}
    \item The first stage categorizes images as lightly, medium, and heavily corrupted based on an adaptive threshold by comparing the M-ALD of processed pixels with the upper and lower MF  of the type-2 fuzzy identifier. 
    \item The second stage eliminates corrupted pixels by computing the appropriate weight using GMF with the mean and variance of the uncorrupted pixels.
\end{enumerate}

The rest of the paper is organized as: An adaptive threshold, and  denoising method is discussed in Section \ref{sec: Proposed Methodology}. Simulation and comparison with different filtering methods are discussed in Section \ref{sec: Results and Discussions}. Finally,   Section \ref{sec: Conclusions} concludes the complete paper.

\section{Preliminaries}
\label{sec: Preliminaries}
%
\subsection{Neighborhood pixel set}
A neighborhood pixel set ${P}^{F}_{ij}$ associated with pixel $p_{ij}\in I$  with \textit{half filter window} of size $F$ is defined as:
\begin{align}\label{eq:R}
\begin{aligned}
{P}^{F}_{ij} = \{\;p_{i+r,j+l} \; \forall\;  r,l \in [-F,F]\;\}
\end{aligned}
\end{align}
where, ${P}^{F}_{ij}$ has a size of $(2F +1)\times(2F + 1)$. 
\subsection{Type-2 fuzzy set} The Type-2 fuzzy set $\tilde{M}_{ij}^{F}$, is characterized by membership function (MF) $\mu_{\tilde{M}_{ij}^{F}}(p_{ij},\mu_{M_{ij}^{F}})$  \cite{c25}  and is defined as:

\begin{align}\label{eq:type 2}
\begin{split}
\tilde{M}_{ij}^{F}= \{\;(p_{ij},\mu_{M_{ij}^{F}}),\;\;\mu_{\tilde{M}_{ij}^{F}}(p_{ij},\mu_{M_{ij}^{F}})\;\; \forall \ p_{ij}\in S,\\  \forall\;\; \mu_{M_{ij}^{F}}\in J_{p_{ij}} \subseteq [0\, 1]\;\}
\end{split}
\end{align}
where $0\leq \mu_{M_{ij}^{F}},  \mu_{\tilde{M}_{ij}^{F}}(p_{ij},\mu_{M_{ij}^{F}}) \leq 1$ and $S$ is the universe of discourse.

\subsection{Mean of k-middle} 
The \textit{mean of $k$-middle} is similar to $\alpha$ trimmed mean as defined in \cite{c14}. Let ${Q}=\{n_{1}, n_{2},\cdots, n_{N}\}$ is an $N$ element set, then the \textit{mean of $k$-middle}, $\Delta_{k}({N})$ is given by
\begin{align}\label{eq:middle}
		\begin{split}
			 \resizebox{0.91\hsize}{!}{%
	$\Delta_{k}({Q})=\left\{
		\begin{array}{@{}ll@{}}
			\frac{1}{2k-1}\sum\limits_{i=h-k+1}^{h+k-1}n_{i},\text{if $N$ is odd}\ (N=2h-1) \\
			\frac{1}{2k}\sum\limits_{i=h-k+1}^{h+k}n_{i},\text{if $N$ is even}\ (N=2h)
		\end{array}\right.$
	}
	\end{split}	
\end{align} 

\section{Proposed Method}
\label{sec: Proposed Methodology}

This section presents two stages to identify and remove the SAP noise. The first stage identifies the pixels as lightly, medium, and heavily corrupted, followed by the second stage, the removal of noisy pixels. The proposed filter is explained in the following subsections.

\subsection{SAP Noise Detection}
\label{subsec: Adaptive Threshold Using Type-2 Fuzzy Logic}
To distinguish the uncorrupted pixels and lightly corrupted pixels, medium corrupted pixels,  from the heavily corrupted pixels, I utilizes the idea of maximum absolute luminance difference (M-ALD) from the approach presented in \cite{h1}.

Initially, a filter window of size $(2F+1)\times(2F+1)$ is chosen  for the calculation of neighborhood pixel set ${\boldsymbol{P}}^{F}_{ij}$.  For point-wise mode of processing, the processed pixel $p_{ij}$ lies at the center of the filter window. The neighborhood pixel set ${\boldsymbol{P}}^{F}_{ij}$ corresponding to processed pixel $p_{ij}$ can be written as
\begin{equation}
\label{processed pixel}
{\boldsymbol{P}}^{F}_{i,j} = \{ p_{i-1, j-1} \}
\end{equation}

The M-ALD of processed pixel $p_{ij}$  with respect to pixel set ${\boldsymbol{P}}^{F}_{ij}$  is defined as 
\begin{align}
\label{ald}
\text{M-ALD } = \max\{|h_{r,l}- p_{i,j}|\} , \; \forall  \;   h_{r,l} \in {\boldsymbol{P}}^{F}_{i,j}  
\end{align}
In a  natural image, it is assumed pixel values change smoothly, and neighborhood pixels tend to have similar pixel values. As we know, when the image is deteriorated by SAP noise, the pixel values may suddenly change to maximum (1) or minimum (0). This approach uses only processed pixels in the filter window to compute the M-ALD. 
After computing M-ALD, in the literature, \cite{h1,h2,h3}  two thresholds are determined to categorize the pixels into three types: lightly, medium, and heavily corrupted. The threshold determined in all these approaches is heuristic and non-deterministic. Due to that, the threshold chosen using these approaches could not be effective in categorizing the pixels. 

Here, an efficient approach is discussed for finding the adaptive threshold to categorize the pixel more efficiently. For determining the  threshold, a fuzzy set $\boldsymbol{P}_{ij}^{F}$ is assumed in the universe of discourse $S$ for all $N$ pixels in the filter window. The MF $\mu_{\boldsymbol{P}_{ij}^{F}}(p_{ij}): {\boldsymbol{P}}^{F}_{ij} \rightarrow [0,1]$ is expressed as primary MF. The primary MFs are chosen as Gaussian MFs (GMFs) with \textit{mean} ($m_{k}^{F}$) with \textit{variance} $(\sigma_{k}^{F})$. These primary MFs help to design the upper MF (UMF) and lower MF (LMF)  to find the threshold for categorizing the pixels as shown in Fig \ref{fig: step of type-2 MFs}.

For every pixel $p_{ij}$  the GMF is defined as
\begin{align}
    \mu_{\boldsymbol{P}_{ij}^{(F,k)}}(p_{ij}) = \exp-\frac{1}{2} \Big( \frac{p_{ij}-m^{F}_{k}}{\sigma^{F}} \Big)^{2}
    \label{eq:primary MFs}
\end{align}


The GMF means ($m_{k}^{F}$) and variance ($\sigma_{k}^{F}$) are calculated as
\vspace{-.1cm}
\begin{align}
	\label{eq:Mean}
	m_{k} = \Delta_{k}({N}^{F}), \;\;& \forall\; \;  k=1,2,3,\dots, h   \\
	\label{eq:Var}
	\sigma_{k} = \Delta_{k}(\Lambda^{F}),\;\; & \forall\; \;  k=1,2,3,\dots, h 
\end{align}
The  parameter $\Lambda^{H}$ is determined using $l_{1}$ norm as


\begin{equation}
    \Lambda^{F}=\epsilon|p_{i,j}-\nu|,\;  \;  \forall \; q,l \in [-F,F]
    \label{eq:omega}
\end{equation}
 where, $\epsilon>1$ is the multiplying factor and $ \nu=\frac{1}{h}\sum_{k=1}^{h}m_{k}.$

The (\ref{eq:Mean})-(\ref{eq:Var}) help to design the upper ($\boldsymbol{ \bar{\mu}}({p}_{n})$) and lower ($ \boldsymbol{\munderbar{\mu}}({p}_{n})$) MFs in a filter window as
\begin{eqnarray}
\begin{aligned}
\label{ULMF}
        \boldsymbol{ \bar{\mu}}({p}_{n})=\left\{
                \begin{array}{ll}
                 \mu_{\boldsymbol{P}_{ij}^{(F,k)}}(m^{F}_{k},\sigma_{k}^{F}),& \text{if}\;p_{n}^{F} < m^{F}_{k}\\
                 \vee(\mu_{\boldsymbol{P}_{ij}^{(F)}}(p_{n})),& \text{if}\;m^{F}_{k}\leq p_{n}^{F} \leq m^{F}_{k^\ast}, k \neq k^\ast\\
              \mu_{\boldsymbol{P}_{ij}^{(F,2)}}(m^{F}_{k},\sigma_{k}^{F}), & \text{if}\; p_{n}^{F} > m^{F}_{k^\ast} \\
                \end{array}
              \right. \\    
        \boldsymbol{ \munderbar{\mu}}({p}_{n})=\left\{
                \begin{array}{ll}
                    \mu_{\boldsymbol{P}_{ij}^{(F,k^\ast)}}(m^{F}_{k^\ast},\sigma_{k}^{F}),\;\;& \text{if}\;p^{F}_{n} \leq \frac{m^{F}_{k}+m^{F}_{k^\ast}}{2},k \neq k^\ast  \\
                     \mu_{\boldsymbol{P}_{ij}^{(F,k)}}(m^{F}_{k},\sigma_{k}^{F}),& \text{if}\; p^F_{n}  > \frac{m^{F}_{k}+m^{F}_{k^\ast}}{2}, k \neq k^\ast
                \end{array}
              \right. 
\end{aligned}
\end{eqnarray}

The  $m_{k}^{F}$ and $\sigma_{k}^{F}$ of GMFs are determined using (\ref{eq:Mean}) and (\ref{eq:Var}) in respective filter window and are shown in Fig. \ref{fig: step of type-2 MFs}. 

Let us define a matrix $\boldsymbol{\tilde{\mu}}$ consisting of UMF and LMF in the filter window. $\boldsymbol{\tilde{\mu}}$  has distinct membership values corresponding to  pixels in the filter window, and it is written as
\begin{align}
\label{eq:PI}
\boldsymbol{\tilde{\mu}} =  
 \begin{bmatrix}
\boldsymbol{\bar{\mu}}({p}_{n})  \\
\boldsymbol{ \munderbar{\mu}}({p}_{n})\\
\end{bmatrix} 
\end{align}
where $\boldsymbol{\bar{\mu}}({p}_{n}) $  and $\boldsymbol{\munderbar{\mu}}({p}_{n})$ having the UMF and LMF values from  $n = 1, 2,\cdots, N$.

A column and row-wise S-norm (max) and a column-wise T-norm (min) followed by S-norm  operation is performed on the matrix $\boldsymbol{\tilde{\mu}}$ to get the upper and lower threshold $T_{h}$ and $T_{l}$  as shown in Fig. \ref{fig: step of type-2 MFs}   and they are defined as
\begin{equation}
\label{eq:Threshold}
T_{h}=\vee(\vee(\boldsymbol{\tilde{\mu}})); \;\; T_{l}=\vee(\wedge(\boldsymbol{\tilde{\mu}}))\;\; \forall \;\; q,l \in [-F,F] 
\end{equation}
where $\wedge$ and $\vee$  are the min and max operators, respectively.

After computing the M-ALD and two adaptive thresholds, a fuzzy flag $f_{ij}$ is defined to categories the pixels into three different types: lightly corrupted pixels, medium corrupted pixels, and heavily corrupted pixels as follows: 
 \begin{figure}
	\centering
	\includegraphics[width=0.62\linewidth]{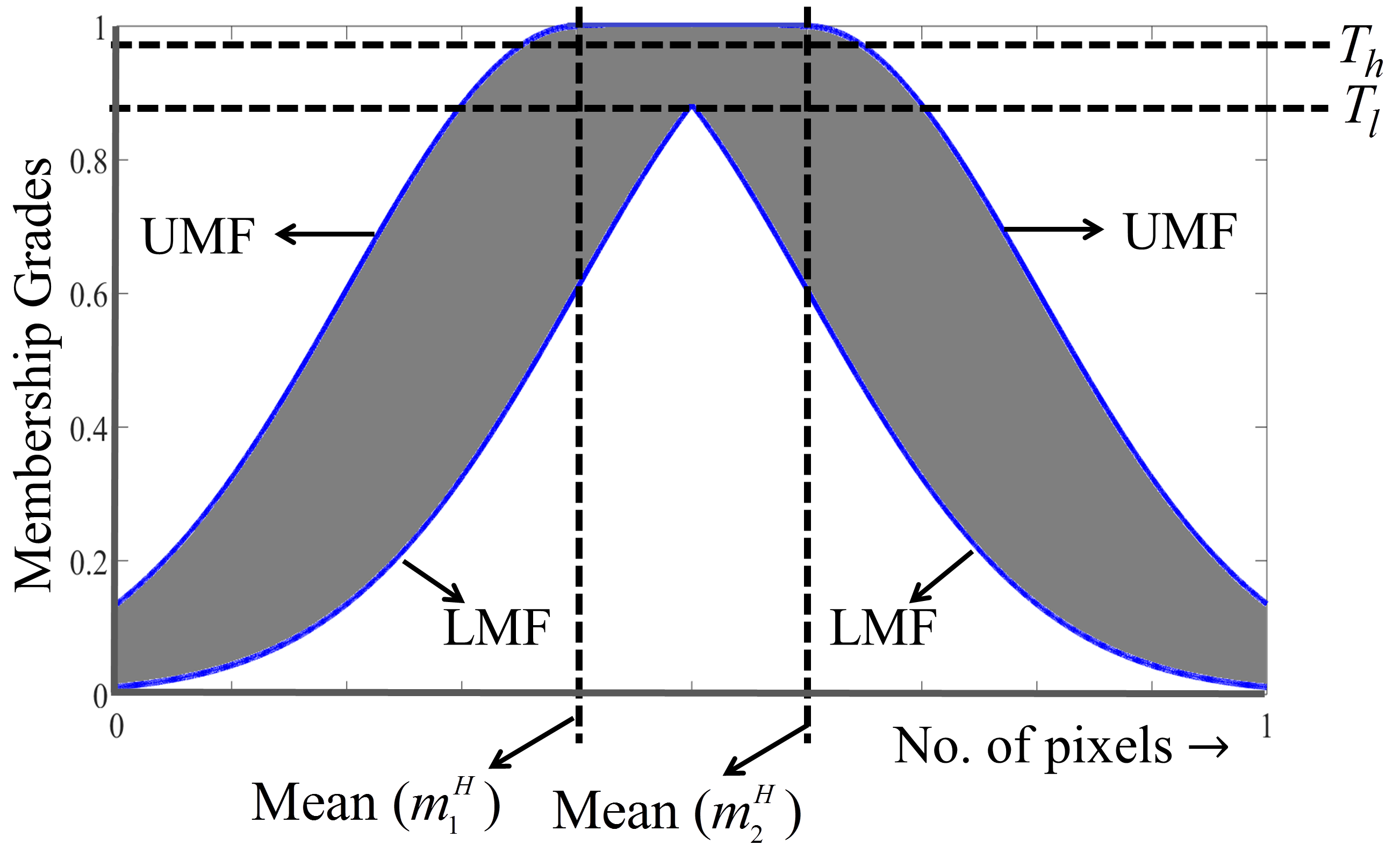}
	\caption{Adaptive threshold using UMF and LMF.}
	\label{fig: step of type-2 MFs}
\end{figure}

\begin{align}
    f_{i j}=\left\{
                \begin{array}{ll}
                  T_{l},  &   \text{if}\; \;  \text{M-ALD} \leq T_{l}   \\
                   \frac{\text{M-ALD}-T_{l}}{T_{h}-T_{1}},  &  \text{if}\; \; T_{l}\leq\text{M-ALD} \leq T_{h} \\ 
                   T_{h},  &   \text{if}\; \;  \text{M-ALD} \geq T_{h}
                \end{array}
              \right.
\label{eq:fuzzy flag}
\end{align}

where, $f_{ij}=T_{l}$ means $p_{ij}$ is lightly corrupted pixel, $0\leq f_{ij}\leq 1$ shows $p_{ij}$ is a medium corrupted pixels and $f_{ij} =T_{h}$ implies $p_{ij}$ is a heavily corrupted pixel.

\subsection{ SAP Noise Removal}
\label{subsec: Denoising using Type-1 Fuzzy Logic}

In this stage, the pixels detected as lightly, mediumly, and heavily corrupted pixels are denoised in the filter window. A set of good pixels $G$ is selected in the filter window for denoising. These good pixels are assumed to be a fuzzy set with Gaussian MF (GMF), and membership values act as a weight assigned to each good pixel. The \textit{mean} ($\nu_G$)  and  \textit{variance} $(\sigma)$ of GMF are computed using $l_{1}$ norm of the good pixels as defined by  (\ref{eq:8}) and (\ref{eq:8varp}). 

The $\nu$ and $\sigma$ of good pixels are computed as
\begin{equation}
    \label{eq:8}
   \nu=\frac{1}{n}\sum_{i=1}^{n}g_{i}\;;\;\;\;\;  \sigma=\frac{1}{n}\sum_{i=1}^{n}\Omega_{{i}}  
\end{equation}
where $n$ is the total number of good pixels and $g_i$ is the $i^{th}$ good pixels. The  parameter $\Omega_{i}$ is determined with $l_{1}$ norm as
\begin{equation}
    \label{eq:8varp}
    \Omega_{{i}}=\epsilon |G-\nu| \;  \; \; \forall \;\; i \in G
\end{equation}

The  GMF of the good pixels is computed as
\begin{equation}
    \label{eq:primary MFG}
    w_{G}(g_{i})	=  \exp-\frac{1}{2} \Big(\frac{g_{i}-\nu}{\sigma} \Big)^{2} 
\end{equation}

The denoised pixel $p^{new}$ corresponding to noisy pixels are evaluated as

\begin{equation}
    \label{eq:10}
    p^{new}=\frac{\sum\limits_{\forall g_{i}\in G}\textit{w}_{i}g_{i}}{W};\;\;\; W=\sum_{i=1}^{r}\textit{w}_{i}
\end{equation}
 where $\textit{w}_{i}\in w_{G}$ be the weight corresponding to the $i^{th}$ good pixel and $W$ is the normalizing weight.

The original pixel in the filter window is estimated using fuzzy flag $f_{i,j}$ as follows:

\begin{align}
	\label{actual}
	\hat{p}_{i,j} =  (1- f_{i,j})p_{i,j} + f_{i,j}p^{new}
\end{align}

where $\hat{p}_{i,j}$  is the estimated pixel of the original pixel $p_{i,j}$ in the filter window. 

\begin{algorithm}
	\caption{AT-2FF}
	\label{alg: Impulse noise}
	\begin{algorithmic}[1]
		\For{all pixels $p_{n} \in I (\text{Image})$}
		\If{$p_{n}\notin\{0,1\}$} retain $p_{n}$ \textbf{continue}
		\EndIf
		\While{$p_{n}\in\{0,1\}$}
		\State \textbf{initialize} $F=1$ for small noise levels and $F=2$ higher noise levels  
		\State Compute M-ALD using (\ref{ald})
		\State Compute $\boldsymbol{\tilde{\mu}}$ using (\ref{eq:PI})
		\State Compute  $T_h$ and $T_l$ using (\ref{eq:Threshold})
		\State Compute  $f_{ij}$ using (\ref{eq:fuzzy flag})
		\If{$mean(\boldsymbol{\tilde{\mu}}(p_{n}))\geq T_{h}$}  retain $p_{n}$  \textbf{break}
		\EndIf
		\If{$\sigma\leq \epsilon $} $p_{n}=\nu$ \textbf{break}
		\EndIf
		\State Compute $ G^{F}_{ij}$ using
		\State $g = |G^{F}_{ij}|$ (number of good pixels)
		\If{$g<1$} $F=F+1$  \textbf{continue}
		\EndIf
		\State Compute $p^{new}$  using (\ref{eq:10})  
		\State Compute denoised pixels using (\ref{actual}) \textbf{break}
		\State \textbf{end while}
		\\
		\textbf{end for}
		\EndWhile
		\EndFor
	\end{algorithmic}
\end{algorithm}

\begin{table*}
	\centering 
	\caption{\textsc{ \small 	
			Comparison of Performance with Six State-of-the-Art Methods (In Term of PSNR (In dB))}}
	\label{tab: PSNR Comparison}
	\begin{tabular}{||c||c|| c|| c||c|| c|| c|| c|| c||}
		\hline 
		{\multirow{1}{*}{\makecell{Input Images}}} &
		{\multirow{1}{*}{\makecell{SAP Noise (in \%)}}}  &
		{\multirow{1}{*}{\makecell{MF \cite{MF}}}} &
		{\multirow{1}{*}{\makecell{TMF \cite{TMF}}}} &
		{\multirow{1}{*}{\makecell{AWMF \cite{AWMF}}}} &
		{\multirow{1}{*}{\makecell{WNNM \cite{WNNM}}}} & 
		{\multirow{1}{*}{\makecell{FRFCM \cite{FRFCM}}}} &
		{\multirow{1}{*}{\makecell{T-2FFMC \cite{T-2FFMC}}}} &
		\multirow{1}{*}{\makecell{AT-2FF}}  \\ \cline{1-9} 
		\multicolumn{1}{ ||c|| }{\multirow{5}{*}{Windows} }& 10 & 35.07 & 24.54  & 24.51 & 23.98 & 25.19 & 33.73  &  \textbf{43.10}\\ 	\cline{2-9} 
		\multicolumn{1}{ ||c|| }{} & 30 & 23.96 & 23.76  & 24.00  & 21.97 & 24.61 &  33.20 & \textbf{35.95 }\\ \cline{2-9}
		\multicolumn{1}{ ||c|| }{} & 50 & 15.52  & 17.00  & 23.52  & 20.11  & 23.17   & 32.17 & \textbf{32.21}\\ \cline{2-9} 
		\multicolumn{1}{ ||c|| }{} & 70 & 10.33 & 12.01  & 23.52 & 19.05 & 18.20    & 30.08  &  29.28\\ \cline{2-9} 
		\multicolumn{1}{ ||c|| }{} & 90 & 6.94 & 6.36 & 23.52  & 15.94  & 6.31  & 23.51  & \textbf{24.62}\\ \hline 
		\multicolumn{1}{ ||c|| }{\multirow{5}{*}{Peppers} }& 10 & 33.27  & 20.45  & 20.44  & 24.03  & 22.64 &  32.36  &  \textbf{43.04} \\ 	\cline{2-9} 
		\multicolumn{1}{ ||c|| }{} & 30 & 23.23    & 20.32 & 20.24  & 21.17  & 22.25 &  30.04  &  \textbf{36.18} \\ \cline{2-9}  
		\multicolumn{1}{ ||c|| }{} & 50 & 15.15 & 16.78 & 19.14 & 18.49 & 20.98 &  27.14  & \textbf{29.59} \\ \cline{2-9} 
		\multicolumn{1}{ ||c|| }{} & 70 & 9.94 & 9.54 & 13.56 & 16.00  & 15.78 &  22.97  & \textbf{24.08} \\ \cline{2-9}  
		\multicolumn{1}{ ||c|| }{} & 90 & 6.57 & 6.38 & 7.34 & 13.14  & 6.24 &  17.61  & \textbf{19.90} \\ \hline 
		\multicolumn{1}{ ||c|| }{\multirow{5}{*}{Mountains} }& 10 & 37.10 & 24.52 & 24.53 & 30.43  & 21.23  & 35.54  &  \textbf{42.55} \\ 	\cline{2-9} 
		\multicolumn{1}{ ||c|| }{} & 30 & 23.78 & 24.31 & 24.39 & 23.36 & 21.19 &  34.33 & \textbf{39.76} \\ \cline{2-9} 
		\multicolumn{1}{ ||c|| }{} & 50 & 15.04  & 19.23  & 22.26  & 18.31  & 20.94 &  33.06  & \textbf{37.65} \\ \cline{2-9} 
		\multicolumn{1}{ ||c|| }{} & 70 & 9.50  & 14.29 & 13.80  & 14.70  & 17.88   & 30.57     & \textbf{36.00} \\ \cline{2-9}  
		\multicolumn{1}{ ||c|| }{} & 90 & 6.15  & 6.30  & 7.09 & 11.55  & 5.95  &  23.82  &   \textbf{30.32}\\ \hline 
		\multicolumn{1}{ ||c|| }{\multirow{5}{*}{Barbara} }& 10 & 28.73 & 19.72 & 19.70 & 23.82  &20.59 &  26.14  & \textbf{37.76} \\ 	\cline{2-9} 
		\multicolumn{1}{ ||c|| }{} & 30 & 21.67  & 19.24  & 19.56 &20.78  & 20.14  & 25.58  &  \textbf{32.10}\\ \cline{2-9} 
		\multicolumn{1}{ ||c|| }{} & 50 & 14.58  & 16.05 & 18.34 & 17.63  & 19.15  & 24.60  & \textbf{27.87}\\ \cline{2-9} 
		\multicolumn{1}{ ||c|| }{} & 70 & 9.56  & 9.50 & 12.95  & 17.80 & 15.22  &  22.67  & \textbf{26.64}\\ \cline{2-9}  
		\multicolumn{1}{ ||c|| }{} & 90 & 6.25 & 6.19  & 12.95  & 12.02  & 6.03 &  18.23  &     \textbf{19.33}\\ \Xhline{2.1\arrayrulewidth}  
		\multicolumn{1}{ ||c|| }{\multirow{5}{*}{Building} }& 10 & 28.42 & 21.91 & 21.94 & 26.64 & 20.97 & 24.80  & \textbf{28.70} \\ 	\cline{2-9} 
		\multicolumn{1}{ ||c|| }{} & 30 & 21.40 & 21.33  & 21.58 & 20.61 & 20.59  &  24.60 & \textbf{27.89}\\ \cline{2-9}  
		\multicolumn{1}{ ||c|| }{} & 50 & 13.73  & 18.22 & 19.85 & 15.72 & 19.70 & 24.04 & \textbf{26.63}\\ \cline{2-9} 
		\multicolumn{1}{ ||c|| }{} & 70 & 8.71 & 10.89 & 12.66 & 12.00 & 14.91 &  23.30  & \textbf{25.28}\\ \cline{2-9} 
		\multicolumn{1}{ ||c|| }{} & 90 & 5.36 & 5.90 & 6.19 & 9.24 & 5.38 &  20.31  & \textbf{22.66}\\ \hline 
	\end{tabular}
\end{table*}

\begin{figure*}
	\centering
	\subfloat{%
		\includegraphics[width=24mm]{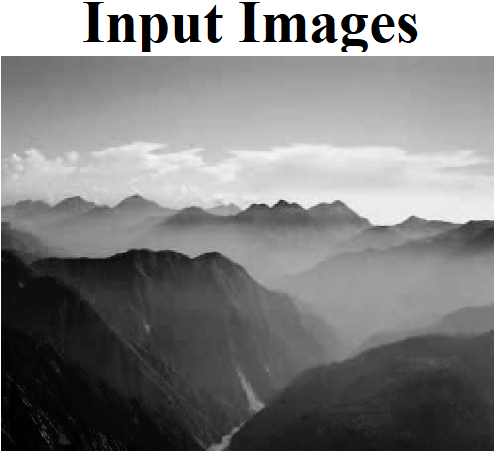}
		\label{subfig: Mountain - Original Image}
	}
	\quad
	\subfloat{%
		\includegraphics[width=24mm]{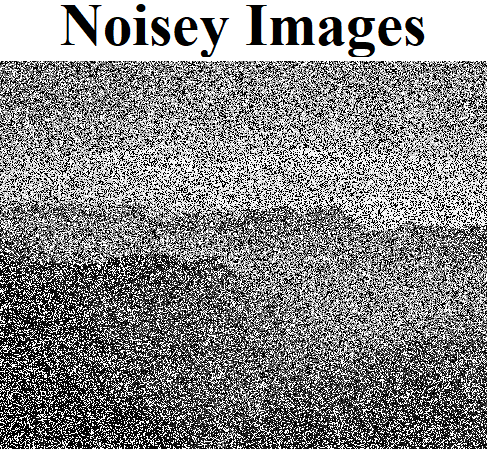}
		\label{subfig: Mountain - at 60 noise}
	}
	\quad
	\subfloat{%
		\includegraphics[width=24mm]{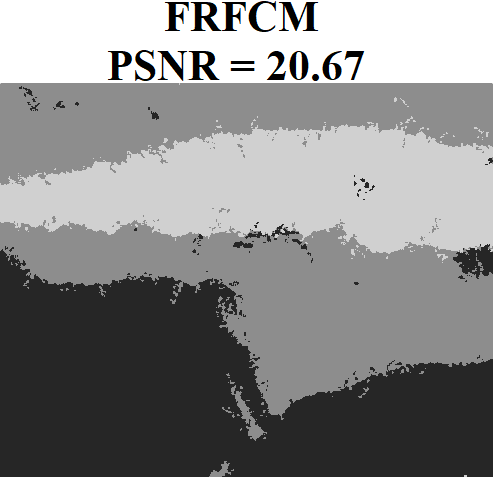}
		\label{subfig: Mountain - filtered image at FRFCM noise}
	}
	\quad
	\subfloat{%
		\includegraphics[width=24mm]{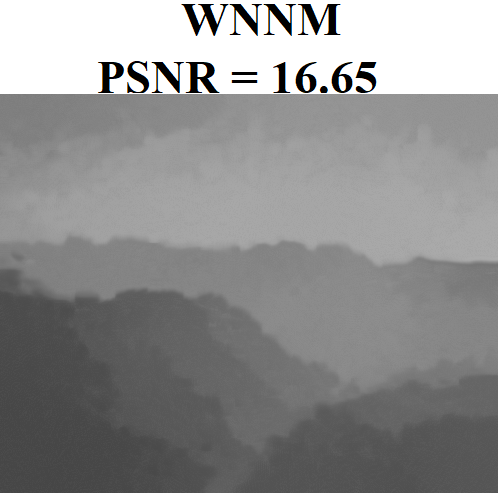}
		\label{subfig: Mountain - filtered image at WNNM noise}
	}
	\quad
	\subfloat{%
		\includegraphics[width=24mm]{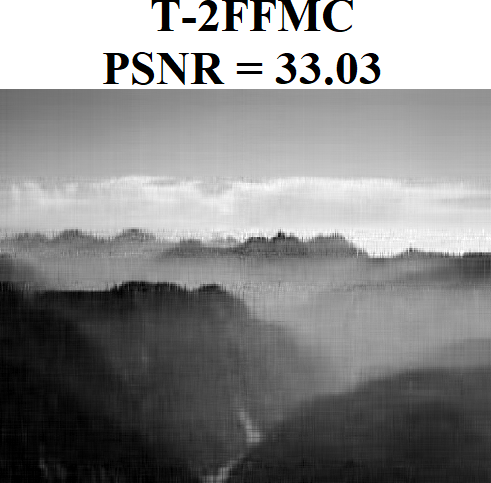}
		\label{subfig: Mountain - filtered image at FMCnoise}
	}
	\quad
	\subfloat{%
		\includegraphics[width=24mm]{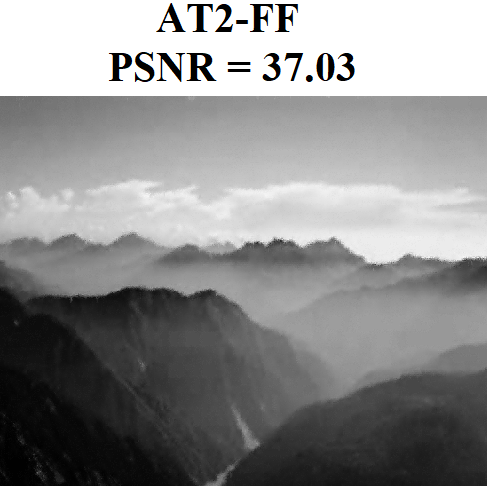}
		\label{subfig: Mountain - filtered image at AT2FF noise}
	}
\end{figure*}

\begin{figure*}
	\centering
	\subfloat{%
		\includegraphics[width=24mm]{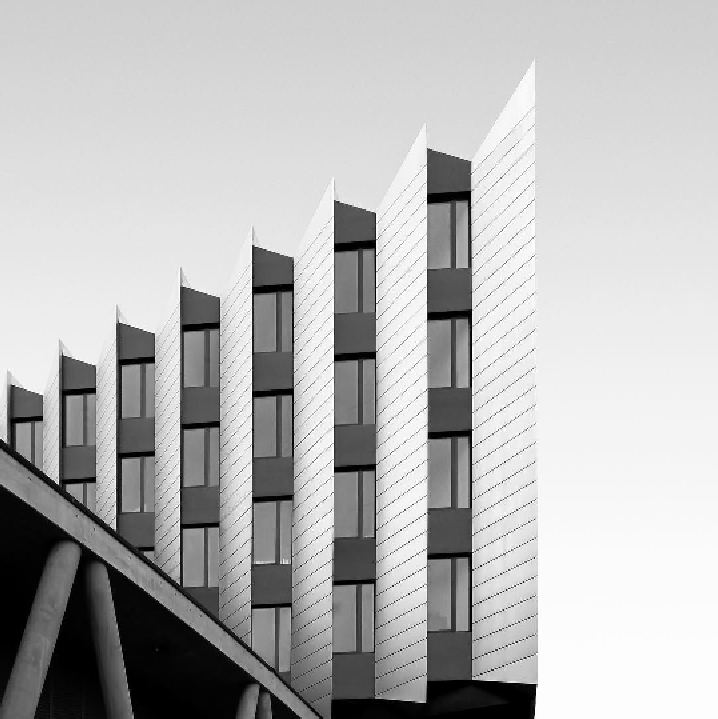}
		\label{subfig: Buildings - Original Image}
	}
	\quad
	\subfloat{%
		\includegraphics[width=24mm]{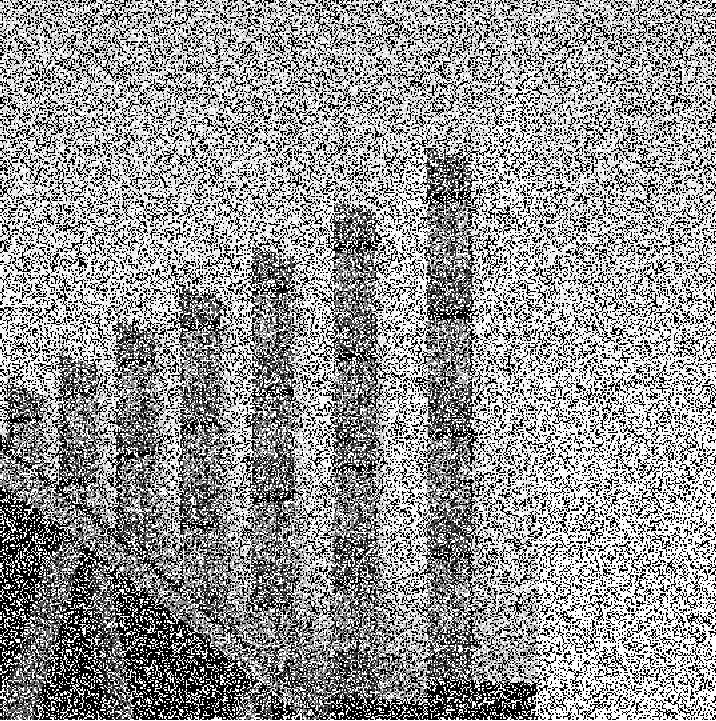}
		\label{subfig: Buildings - at 60 noise}
	}
	\quad
	\subfloat{%
		\includegraphics[width=24mm]{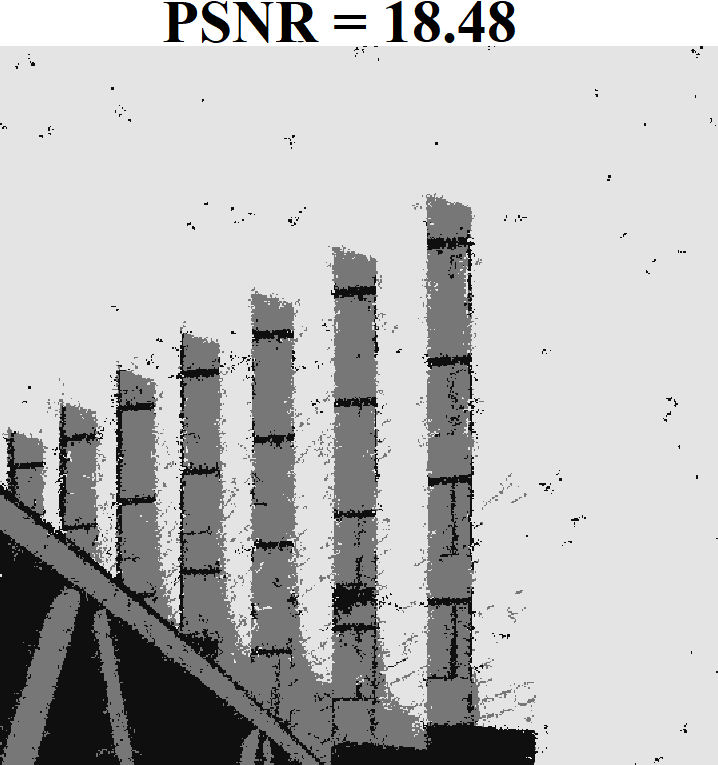}
		\label{subfig: Buildings - filtered image at FRFCM noise}
	}
	\quad
	\subfloat{%
		\includegraphics[width=24mm]{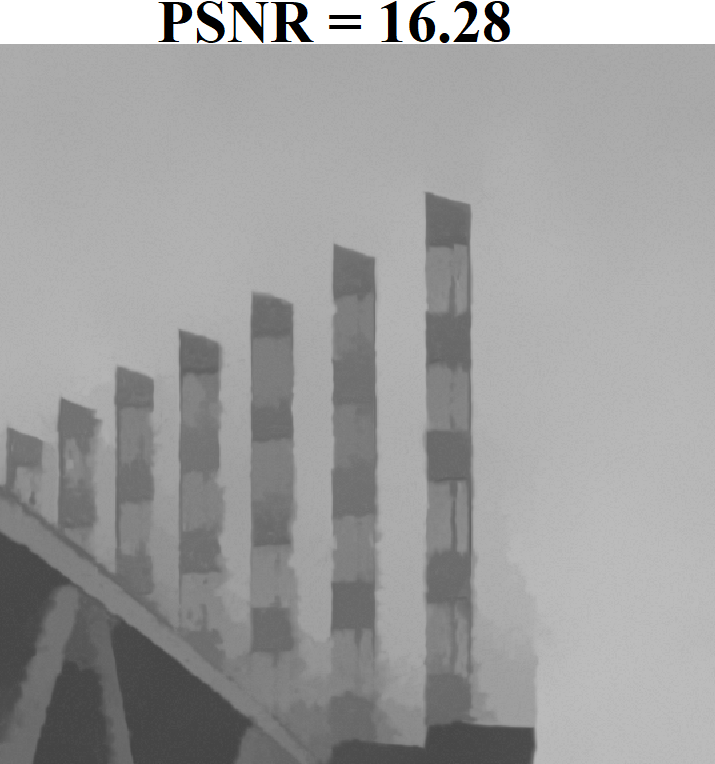}
		\label{subfig: Buildings - filtered image at WNNM noise}
	}
	\quad
	\subfloat{%
		\includegraphics[width=24mm]{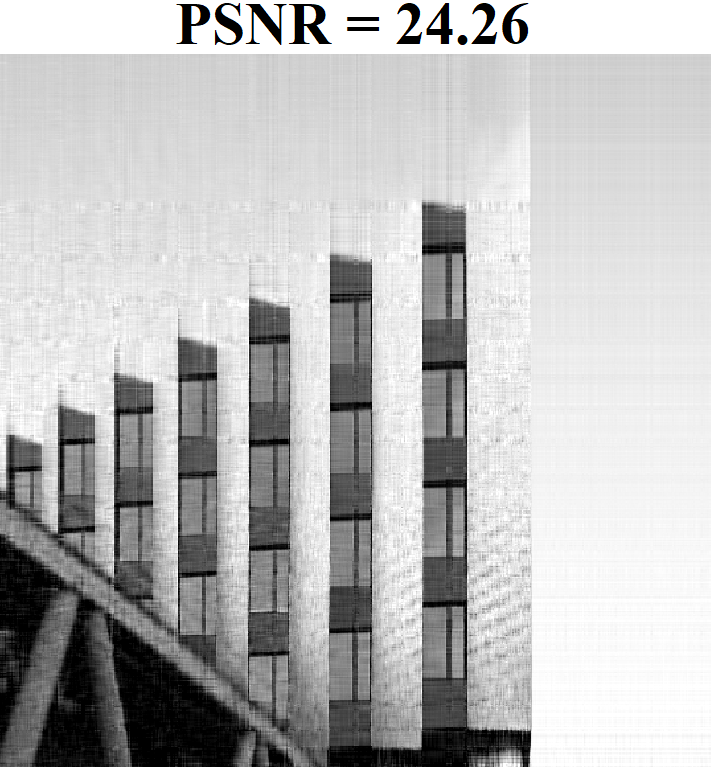}
		\label{subfig: Buildings - filtered image at FMC noise}
	}
	\quad
	\subfloat{%
		\includegraphics[width=24mm]{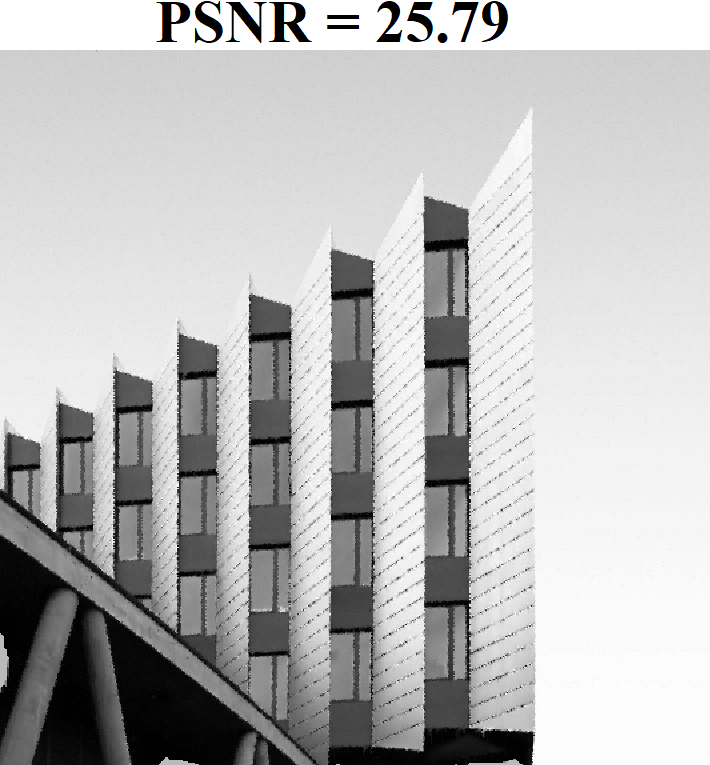}
		\label{subfig: Buildings - filtered image at AT2FF noise}
	}
\end{figure*}
\begin{figure*}
	\centering
	\subfloat{%
		\includegraphics[height=24mm]{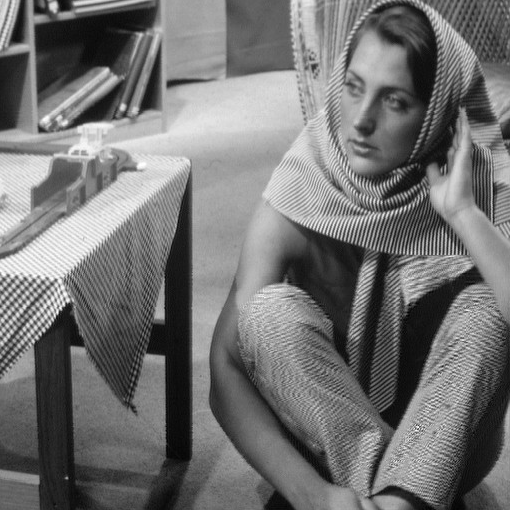}
		\label{subfig: barbara - Original Image}
	}
	\quad
	\subfloat{%
		\includegraphics[height=24mm]{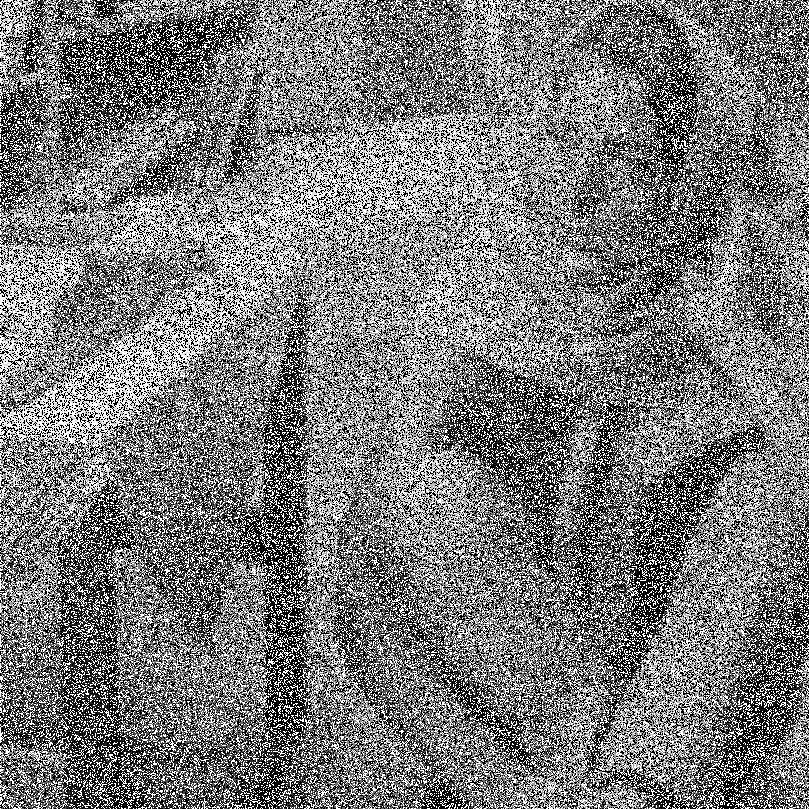}
		\label{subfig: barbara - at 60 noise}
	}
	\quad
	\subfloat{%
		\includegraphics[height=26mm]{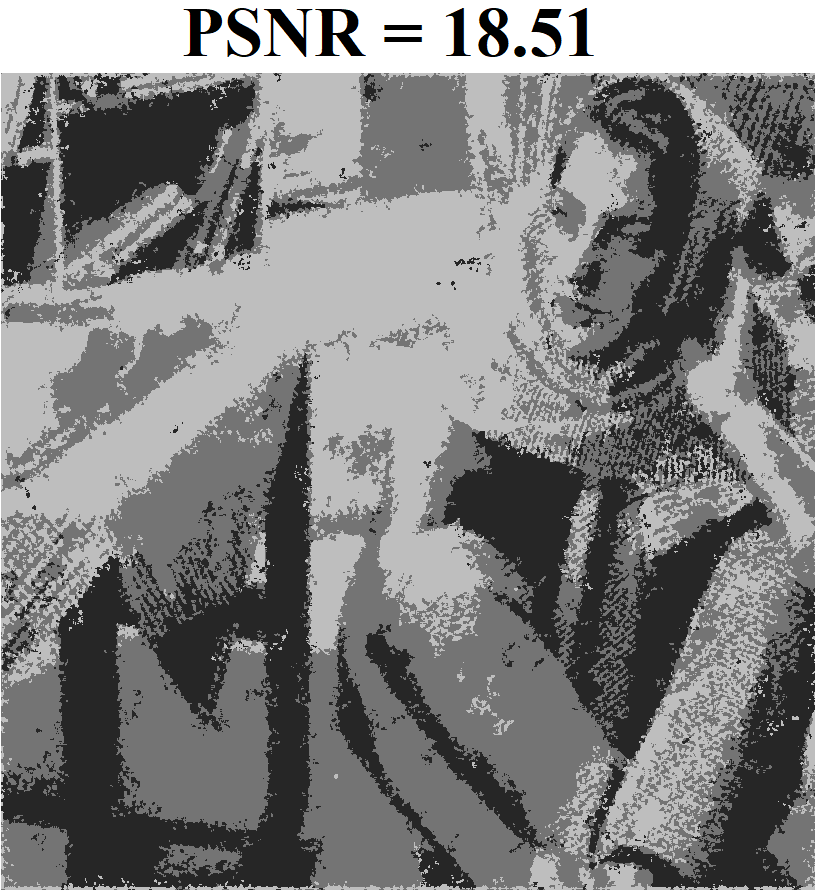}
		\label{subfig: barbara - filtered image at FRFM noise}
	}
	\quad
	\subfloat{%
		\includegraphics[height=26mm]{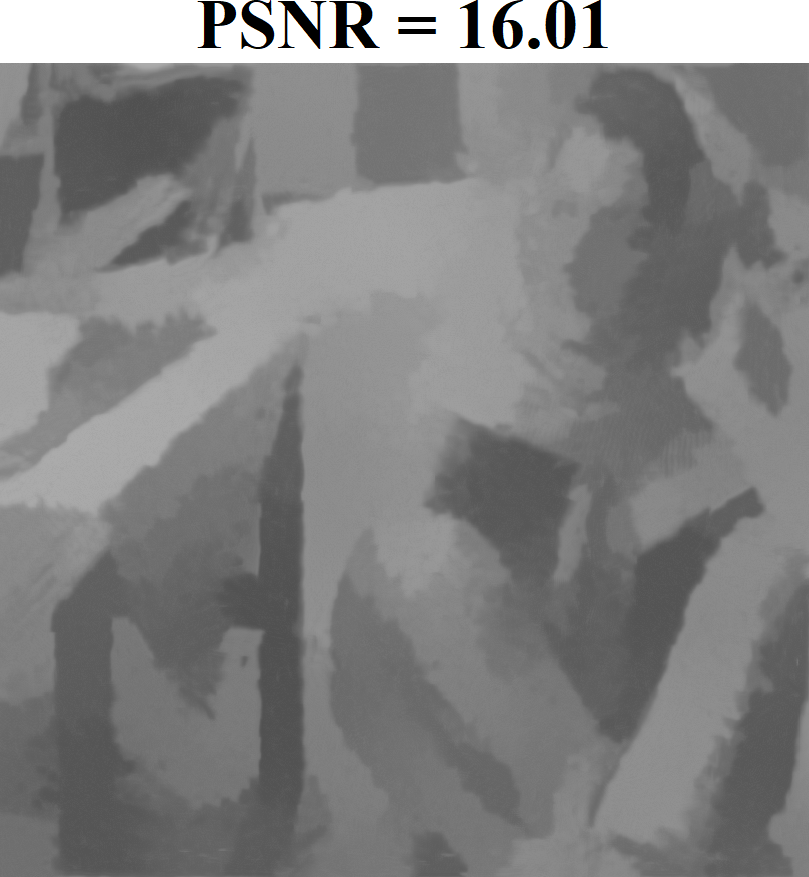}
		\label{subfig: barbara - filtered image at WNNM noise}
	}
	\quad
	\subfloat{%
		\includegraphics[height=26mm]{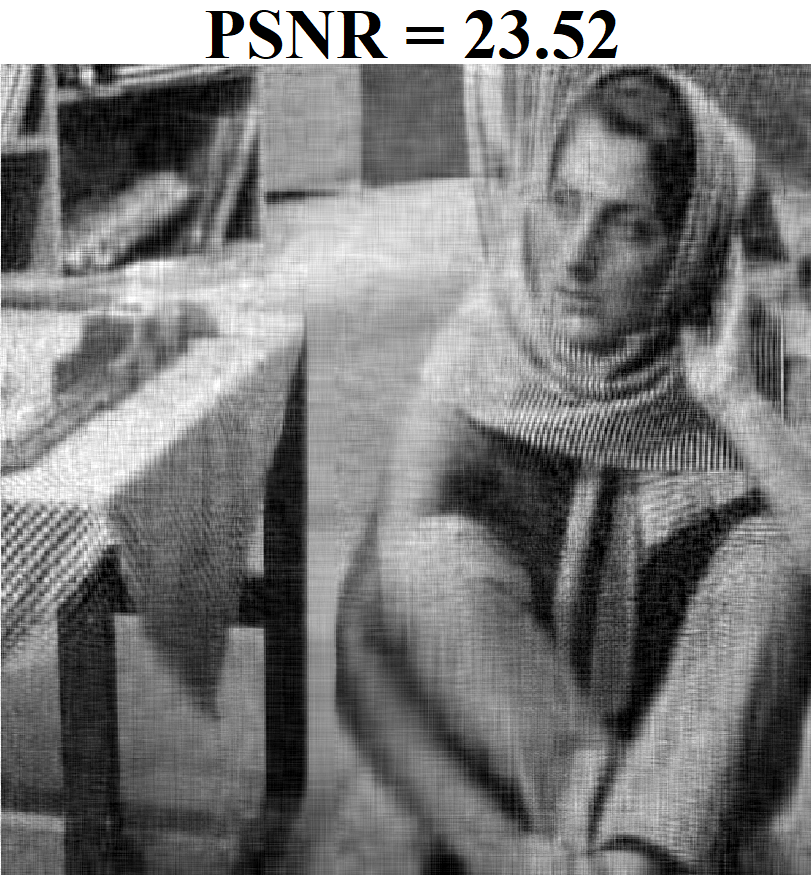}
		\label{subfig: barbara - filtered image at FCM noise}
	}
	\quad
	\subfloat{%
		\includegraphics[height=26mm]{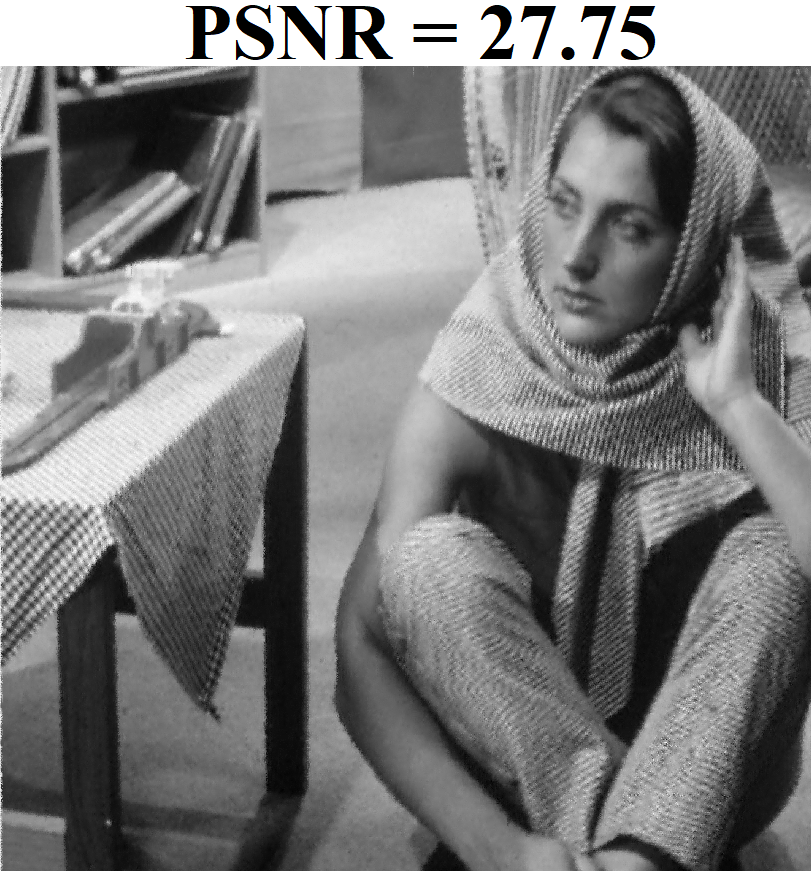}
		\label{subfig: barbara - filtered image at AT2FF noise}
	}
\end{figure*}
\begin{figure*}
	\centering
	\subfloat{%
		\includegraphics[width=24mm]{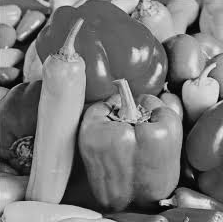}
		\label{subfig: Peppers - Original Image}
	}
	\quad
	\subfloat{%
		\includegraphics[width=24mm]{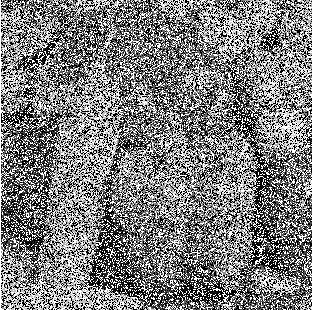}
		\label{subfig: Peppers - at 60 noise}
	}
	\quad
	\subfloat{%
		\includegraphics[width=24mm]{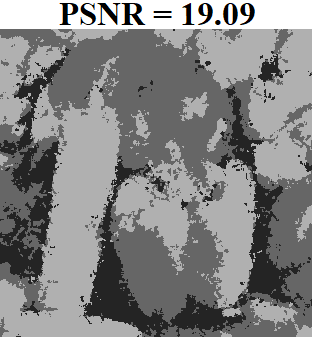}
		\label{subfig: Peppers - filtered image at FRFM noise}
	}
	\quad
	\subfloat{%
		\includegraphics[width=24mm]{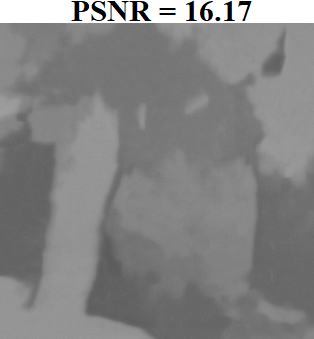}
		\label{subfig: Peppers- filtered image at WNNM noise}
	}
	\quad
	\subfloat{%
		\includegraphics[width=24mm]{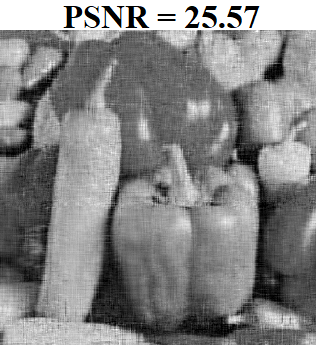}
		\label{subfig: Peppers - filtered image at FMC noise}
	}
	\quad
	\subfloat{%
		\includegraphics[width=24mm]{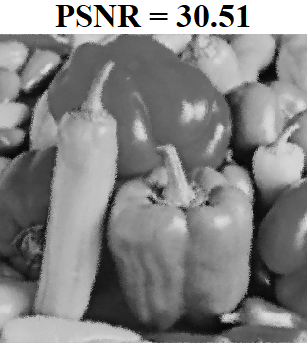}
		\label{subfig: Peppers - filtered image at AT2FF noise}
	}
\end{figure*}
\begin{figure*}
	\centering
	\subfloat{%
		\includegraphics[height=23.7mm]{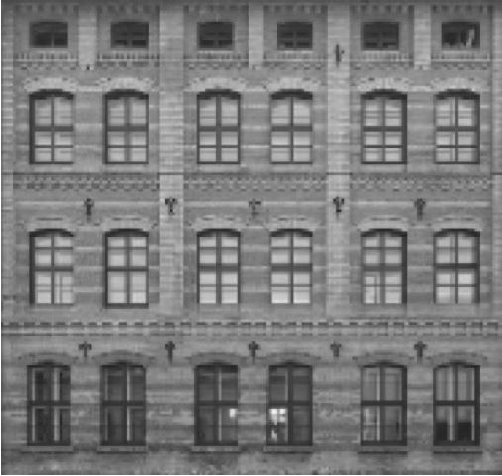}
		\label{subfig: Lena - Original Image}
	}
	\quad
	\subfloat{%
		\includegraphics[height=23.7mm]{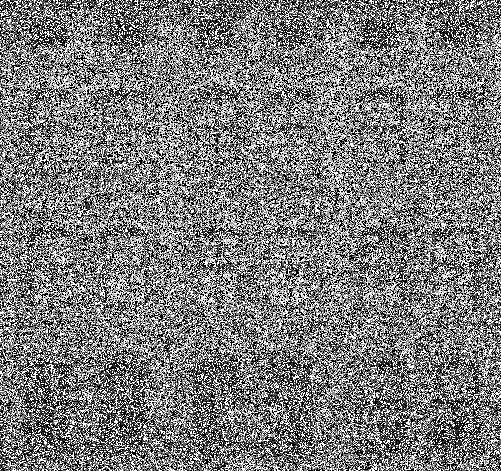}
		\label{subfig: Windows - at 60 noise}
	}
	\quad
	\subfloat{%
		\includegraphics[height=24.2mm]{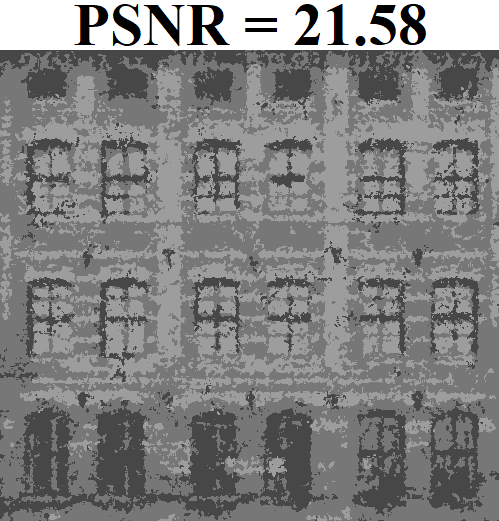}
		\label{subfig: Windows - filtered image at FRFCM noise}
	}
	\quad
	\subfloat{%
		\includegraphics[height=24.2mm]{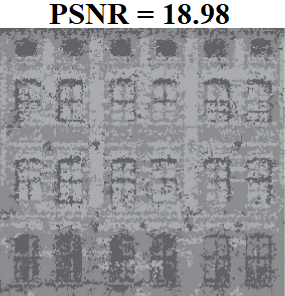}
		\label{subfig: Windows - filtered image at WNNM noise}
	}
	\quad
	\subfloat{%
		\includegraphics[height=24.2mm]{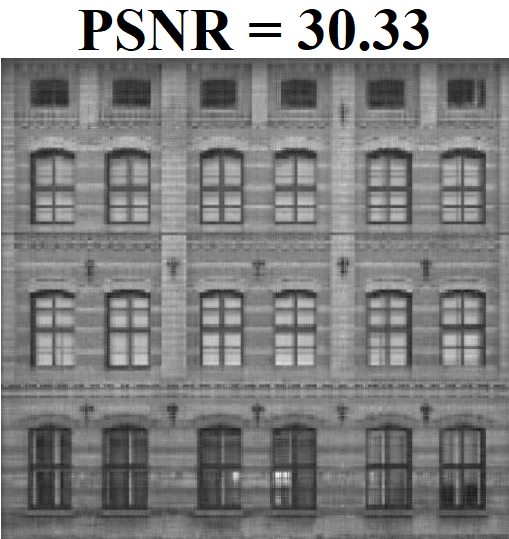}
		\label{subfig: Windows - filtered image at FMC noise}
	}
	\quad
	\subfloat{%
		\includegraphics[height=24.2mm]{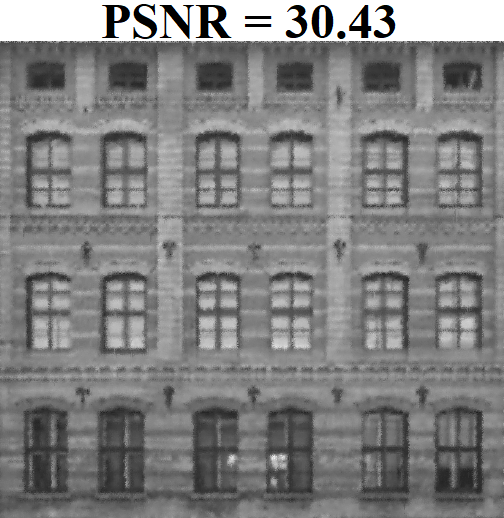}
		\label{subfig: Windows - filtered image at AT2FF noise}
	}
	
	\caption{\small First column: Input images i.e., Mountains, Building, Barbara, Peppers, and Windows. Second column: Noisy images with 60\% SAP noise density. Followed by the de-noising methods:  FRFCM \cite{FRFCM}, WNNM \cite{WNNM},  T-2FFMC \cite{T-2FFMC} and AT-2FF in each columns, respectively.}
\end{figure*}

 \begin{table*}
 	\centering
 	\caption{ \textsc{ \small Comparison of Average Computation Time with Six State-of-the-Art Methods (in second)}}
 	\label{tab: Run Time Comparison}
 	\begin{tabular}{||c|| c|| c|| c|| c|| c|| c||c|| }
 		\hline
 		{\multirow{2}{*}{\makecell{SAP Noise  (in \%)}}}  &
 		{\multirow{2}{*}{\makecell{MF \cite{MF}}}} &
 		{\multirow{2}{*}{\makecell{TMF \cite{TMF}}}} &
 		{\multirow{2}{*}{\makecell{AWMF  \cite{AWMF}}}} &
 		{\multirow{2}{*}{\makecell{WNNM  \cite{WNNM}}}} & 
 		{\multirow{2}{*}{\makecell{FRFCM  \cite{FRFCM}}}} &
 		{\multirow{2}{*}{\makecell{T-2FFMC  \cite{T-2FFMC}}}} &
 		{\multirow{2}{*}{\makecell{AT-2FF}}} \\
 		& &  &  &  &  &  &        \\	 \hline
 		\multicolumn{1}{||c||}{\multirow{1}{*}{10} }& 1.30 & 1.51   & 733.26  &  1866.96 & 2.00 &  947.78  &     \textbf{6.34}\\ \hline
 		\multicolumn{1}{||c||}{\multirow{1}{*}{30} } & 1.30  & 1.51  &  870.27 & 5732.95  & 2.05 &  2514.45 &   \textbf{14.26}\\ \hline
 		\multicolumn{1}{||c||}{\multirow{1}{*}{50} } & 1.30   & 1.51  & 863.46  &9237.71   & 2.10 & 3744.79  &   \textbf{18.77}\\ \hline
 		\multicolumn{1}{||c||}{\multirow{1}{*}{70} } & 1.30  & 1.51 & 699.60 & 12207.72 & 2.15 & 4546.15 &   \textbf{18.18}\\ 
 		\hline 
 		\multicolumn{1}{||c||}{\multirow{1}{*}{90} } & 1.30  & 1.51 & 650.90 & 15011.72 & 2.23 & 4929.35 &   \textbf{17.23}\\ 
 		\hline 
 	\end{tabular}
 \end{table*}

\begin{figure*}
	\centering
	\subfloat{%
		\includegraphics[height=23.5mm]{Pictures/mountain_input.png}
		\label{subfig: Lena - Original Image}
	}
	\quad
	\subfloat{%
		\includegraphics[height=23mm]{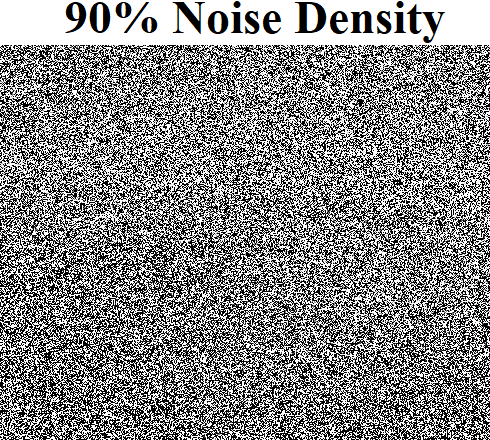}
		\label{subfig: Lena - at 97 noise}
	}
	\quad
	\subfloat{%
		\includegraphics[height=24.5mm]{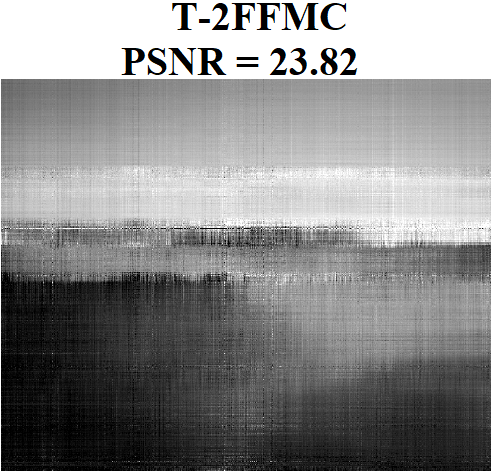}
		\label{subfig: Lena - filtered image at 97 noise}
	}
	\quad
	\subfloat{%
		\includegraphics[height=24.5mm]{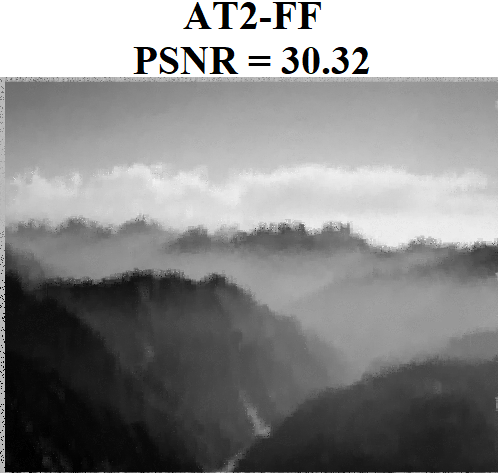}
		\label{subfig: Boat - Original Image}
	}
	\quad
	\subfloat{%
		\includegraphics[height=23mm]{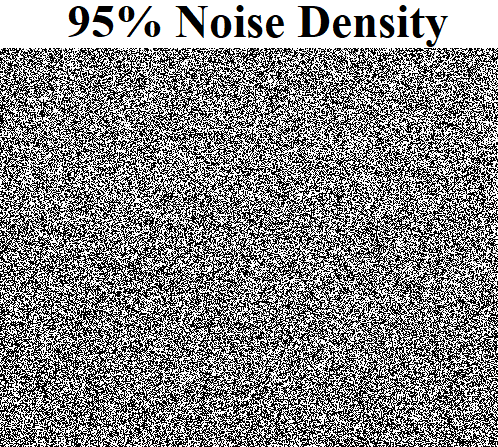}
		\label{subfig: Boat - at 97 noise}
	}
	\quad
	\subfloat{%
		\includegraphics[height=24.5mm]{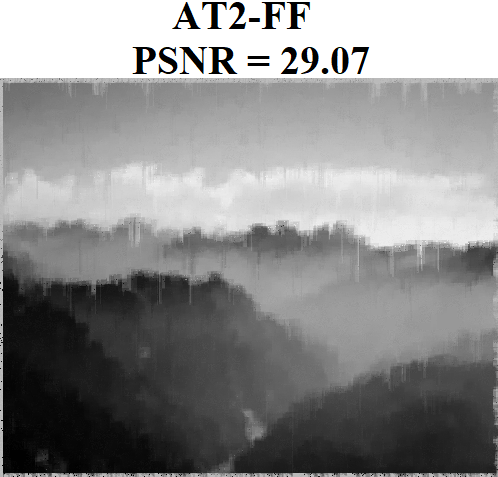}
		\label{subfig: Boat - filtered image at 99 noise}
	}
	\caption{(a)  Original Image (b) mountain - with 90\% SAP noise density (c) Mountain - restored image with T2-FFMC (d) Mountain - restored image with AT2-FF (e) Mountain - with 95\% SAP noise density (f) Mountain - restored image with AT-2FF.}
\end{figure*}

\section{Simulation Results}
\label{sec: Results and Discussions}
The performance of the AT-2FF filter is examined quantitatively and qualitatively in terms of PSNR  at different SAP noise levels on five standard grayscale images, i.e., Mountains ($500 \times 400$), Building ($1080 \times 1080$), Barbara ($817 \times 817$), Peppers ($317 \times 316$), and Windows ($508 \times 481$), respectively. 

The PSNR is defined using recovered image $(\boldsymbol{I}_{r})$  with respect to original image $(\boldsymbol{I}_{o})$ as
\begin{equation}
	\begin{array}{@{}ll@{}}
		\textit{PSNR}(\boldsymbol{I}_{o}, \boldsymbol{I}_{r})= 10 \log_{10} \frac{(255)^{2}}{\frac{1}{XY}\sum_{i,j}(\boldsymbol{I}_{o}(i,j) - \boldsymbol{I}_{r}(i,j))^{2}}
	\end{array}
\end{equation}

where 255 is the maximum pixel intensity for 8-bit images.

The effectiveness of the AT-2FF filter has been compared quantitatively with six different filters, i.e., MF \cite{MF}, TMF \cite{TMF}, AWMF \cite{AWMF}, WNNM \cite{WNNM}, FRFCM \cite{FRFCM}, and T-2FFMC \cite{T-2FFMC}, at noise density levels i.e., 10\%, 30\%, 70\%, and 90\%, are given in Table \ref{tab: PSNR Comparison}. At SAP noise level of 60\%, recovered images of Mountains, Building, Barbara, Peppers, and Windows are shown in Fig. 2 concerning AT-2FF. The proposed filter preserves more image characteristics than other widely used filter approaches. I have also compared the results at higher noise levels, i.e., 90\% and 95\%. The only two filters, i.e., T-2FFMC and AT-2FF, can filter the noise at 90\% noise level. However, at a 95\%  noise level, only the present filter (AT-2FF) can filter the noise and preserve the image details, as shown in Fig. 3. The current filter is more effective at both higher and lower noise levels due to the adaptive threshold chosen with the help of M-ALD and denoised pixels are weighted based on the weight obtained by GMF as explained in the methods. In most of the approaches, the denoised pixels are equally weighted. Because of that, they are unable to provide essential image features at higher noise.  

The computation time of the AT-2FF is also compared with state-of-the-art filters at noise density levels, i.e., 10\%, 30\%, 70\%, and 90\%, which are provided in Table \ref{tab: Run Time Comparison}. The computation time of AT-2FF increases as I increase the noise density levels; however, at higher noise density levels, the filter window size ($F=2$) is increased, which again reduces the computation time as shown in the table for the noise densities 70\% and 90\%. All the experiments are performed on MATLAB 2023a with a 3.40 GHz Core i9 CPU and 128 GB RAM under a 64-bit Linux operating system.

\section{Conclusion}
\label{sec: Conclusions}
This paper presents an adaptive type-2 fuzzy filter for de-noising images corrupted with salt-and-pepper noise (AT-2FF). It includes two stages: In the first stage, an adaptive threshold detects the noisy pixel based on M$-$ALD in a filter window. The detected pixel is denoised by computing the appropriate weight using GMF with the mean and variance of the uncorrupted pixels in the filter window. The AT-2FF filter is experimented with five standard grayscale images at noise levels of 10\%, 30\%, 50\%, 70\%, and 90\% and compared with six different filtering methods. The filtered images at 60\% 90\% and 95\% are shown in Figs. 2 and 3 and computation time \ref{tab: Run Time Comparison} which shows that the present filter is very effective for the image processing for restoring the original image to speed up the image processing task, such as image segmentation, edge detection, and object recognition at higher noise levels.

\bibliographystyle{IEEEtran}
\bibliography{Reference.bib}

\end{document}